\pdfoutput=1

\documentclass[11pt]{article}

\usepackage[preprint]{acl}

\usepackage{times}
\usepackage{latexsym}

\usepackage[T1]{fontenc}

\usepackage[utf8]{inputenc}

\usepackage{microtype}

\usepackage{inconsolata}

\usepackage{graphicx}
\usepackage{xspace}
\usepackage{booktabs}
\usepackage{boldline}
\usepackage{multirow}
\usepackage{cleveref}
\usepackage{caption}
\usepackage{subcaption}
\usepackage{listings}
\usepackage{wrapfig}
\usepackage{enumitem}
\usepackage[section]{placeins}
\usepackage{soul}
\usepackage{booktabs} 
\usepackage{graphicx}
\usepackage{threeparttable}
\usepackage{amssymb}

\newcommand\blfootnote[1]{%
  \begingroup
  \renewcommand\thefootnote{}\footnote{#1}%
  \addtocounter{footnote}{-1}%
  \endgroup
}

\setlist[itemize]{leftmargin=*}
\title{{Learning to Explore and Select for Coverage-Conditioned Retrieval-Augmented Generation}}

\author{
 \textbf{Takyoung Kim\textsuperscript{1,\textdagger}}\quad
 \textbf{Kyungjae Lee\textsuperscript{2}}\quad
 \textbf{Young Rok Jang\textsuperscript{2}} \\
 \textbf{Ji Yong Cho\textsuperscript{2,3}}\quad
 \textbf{Gangwoo Kim\textsuperscript{4,\textdagger}}\quad
 \textbf{Minseok Cho\textsuperscript{2}}\quad
 \textbf{Moontae Lee\textsuperscript{2,5}}
\\
\\
 \textsuperscript{1}University of Illinois Urbana-Champaign\quad
 \textsuperscript{2}LG AI Research \\
 \textsuperscript{3}Cornell University\quad
 \textsuperscript{4}Korea University\quad
 \textsuperscript{5}University of Illinois Chicago
\\
 {
    \href{mailto:tk30@illinois.edu}{\texttt{tk30@illinois.edu}} \qquad
    \href{mailto:moontae.lee@lgresearch.ai}{\texttt{moontae.lee@lgresearch.ai}}
 }
}

\newcommand{\ie}{\textit{i}.\textit{e}.}
\newcommand{\eg}{\textit{e}.\textit{g}.}
\newcommand{\model}{\textsc{QPlanner}\xspace}
\newcommand{\dataset}{\textsc{QTree}\xspace}

\newcommand{\incl}{\textsc{Inclusion}\xspace}
\newcommand{\excl}{\textsc{Exclusion}\xspace}

\newcommand{\cc}{\textsc{C$^2$}\xspace}

\begin{document}
\maketitle

\blfootnote{\textsuperscript{\textdagger}Work done as a research intern at LG AI Research.}

\begin{abstract}
Interactions with large language models (LLMs) often yield long and detailed responses, leveraging both parametric knowledge and retrieval-augmented generation (RAG). While these responses can provide rich insights, they often include redundant or less engaging content not aligned with user interests. This issue becomes apparent when users specify particular subtopics to include or exclude -- termed \textbf{coverage-conditioned (\cc) queries} -- as LLMs often struggle to provide tailored responses. To address this challenge, we investigate the role of \textit{query outlines}, sequences of subqueries designed to guide LLMs in generating responses that meet specific user requirements. To systematically create and evaluate these outlines, we introduce \textbf{\dataset}, a dataset of 10K hierarchical sets of information-seeking subqueries that define structured boundaries for outline creation and evaluation in \cc scenarios\footnote{Our resources are available at \url{https://github.com/youngerous/qtree}.}. Additionally, we develop \textbf{\model}, a 7B language model trained to generate customized outlines within boundaries of \dataset. We evaluate the effectiveness of the generated outlines through automatic and human judgements, focusing on their impact within retrieval-augmented generation (RAG) systems. Experimental results demonstrate that \model, especially when trained with alignment techniques like DPO, generates higher-quality outlines that better fulfill diverse user needs. 
\end{abstract}

\section{Introduction}

\begin{figure}[t!]
    \centering
    \includegraphics[width=\columnwidth]{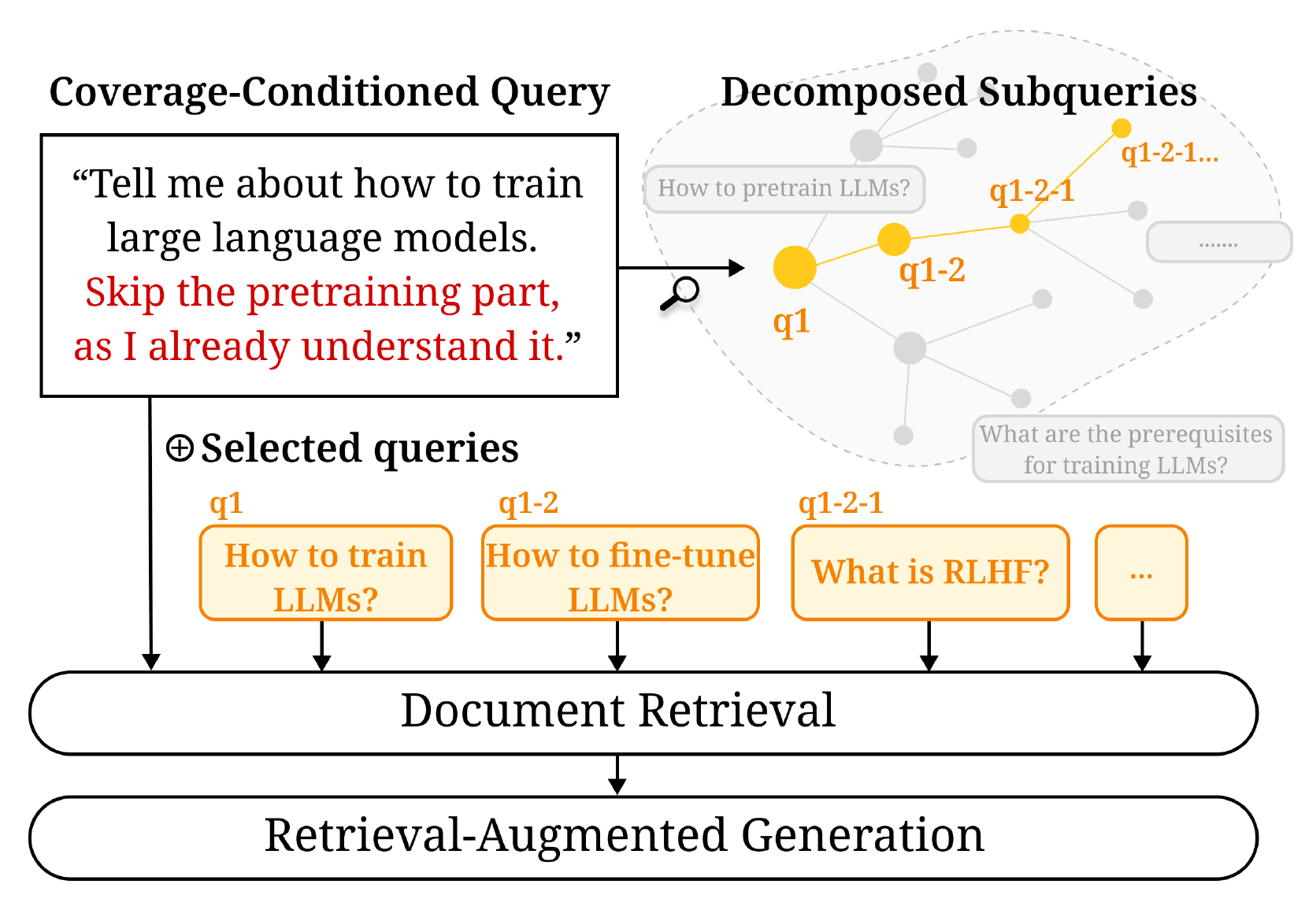}
    \caption{\dataset constrains the range of available outlines for the user's \cc query, and tailored outlines satisfying the requirement of \cc query are selected for RAG downstream tasks.
    }
    \label{fig:teaser}
    \vspace{-3mm}
\end{figure}

Recent advancements of large language models (LLMs) have enabled them to provide long and detailed responses by leveraging their parametric knowledge. As these models improve, human-machine interaction interfaces (\eg, chat and acoustic interfaces) -- which have been studied for a long time~\citep{817450} -- have become more sophisticated, allowing users to request highly specific and personalized information. Proprietary chat services such as ChatGPT~\citep{openai2023gpt4}, Gemini~\citep{team2023gemini}, and BingChat have further accelerated the exploration of personalized information. Additionally, retrieval-augmented generation (RAG) methods are being adopted to enhance the relevance and timeliness of LLM responses by integrating external knowledge.

Despite these advancements, LLMs often struggle with delivering tailored responses when faced with complex user queries. For instance, a user might request LLMs to provide information on \textit{Generative AI}, focusing specifically on its historical context while \textit{excluding} recent trends. Crafting such meticulously composed responses is difficult for LLMs for two reasons: (1) LLMs' long-form outputs can contain innumerable combinations of relevant topics, and (2) there is no established gold standard for long-form text generation~\citep{krishna-etal-2021-hurdles, xu-etal-2022-answer, xu-etal-2023-critical}. Recognizing this, we first define queries that constrain the information coverage of certain topics as \textbf{coverage-conditioned (\cc) queries}, where ``coverage'' refers to the user's intent to instruct LLMs to include or exclude specific subtopics within their responses. These \cc queries especially pose challenges in constructing long-form RAG responses as they require selective document retrieval as well.

To improve LLM responses for users' complex queries, there have been works on \textit{query outlining}, creating sequences of intermediate subtopics to guide long-form responses. Query outlining has been effective in areas like long story generation~\citep{fan-etal-2018-hierarchical, sun-etal-2022-summarize, yang-etal-2022-re3, yang-etal-2023-doc, wang2024weaver, shao2024assisting}. However, generating high-quality outlines that address complex queries like \cc queries remains challenging, as there is no systematic approach for creating and evaluating such outlines. 

With the concepts of \cc query and query outlining in place, we pose two key research questions:

\begin{enumerate}[label=\textit{RQ\arabic*.} , left=1em]
    \item \textit{How can we create and evaluate better outlines for \cc queries?}
    \item \textit{Can these outlines improve RAG systems by serving as search queries and content drafts?}
\end{enumerate}

To address \textit{RQ1}, we present \textbf{\dataset}, a dataset comprising 10K hierarchical sets of information-seeking subqueries (with 39 subqueries in each set) that interpret user queries with diverse perspectives, facilitating the exploration and selection of appropriate outlines for \cc queries. The hierarchies in \dataset are organized according to the abstraction level of the main topic, defining tangible boundaries of available outlines. For example, as illustrated in Figure~\ref{fig:teaser}, hierarchical subtopics related to processes after pretraining (\ie, \textit{Fine-tuning} and \textit{RLHF}) are selected as proper outlines for RAG response among various viewpoints on the topic of \textit{Training LLMs}, following the requirements of the \cc query. In contrast, less relevant subtopics in \dataset (\eg, \textit{Pretraining LLMs}) will not be a desirable outline for the \cc query. By leveraging \dataset, we can systematically create and judge outlines for long-form responses, ensuring that they align with the user’s coverage constraints. 

Regarding \textit{RQ2}, we introduce \textbf{\model}, an autoregressive 7B language model designed to generate tailored outlines within \dataset's hierarchical boundaries. We hypothesize that high-quality outlines aligned with \cc queries can improve both document retrieval and response generation in RAG systems. We also evaluate \model’s performance through both automatic metrics and human judgments, assessing the quality of the generated outlines and their impact on downstream tasks. Experimental results on \cc queries from diverse domains (\ie, Wikipedia and expert domains) demonstrate that training \model with preference alignment further improves both outline quality and overall RAG performance.

Our contributions are summarized as follows:
\vspace{-2mm}
\begin{enumerate}
    \item We present \dataset, a novel dataset of 10K hierarchical subquery sets that define boundaries for available outlines, facilitating the creation and evaluation of better outlines for coverage-conditioned (\cc) queries (addressing \textit{RQ1}).
    \item We introduce \model, an autoregressive language model designed to generate customized outlines that improve document retrieval and content generation in RAG systems (addressing \textit{RQ2}).
    \item We conduct comprehensive evaluations, including automatic metrics and human judgments, to validate the effectiveness of our approach in enhancing outline quality and RAG performance.
\end{enumerate}

\section{Related Work}

\begin{figure*}[ht]
    \centering
    \includegraphics[width=\textwidth]{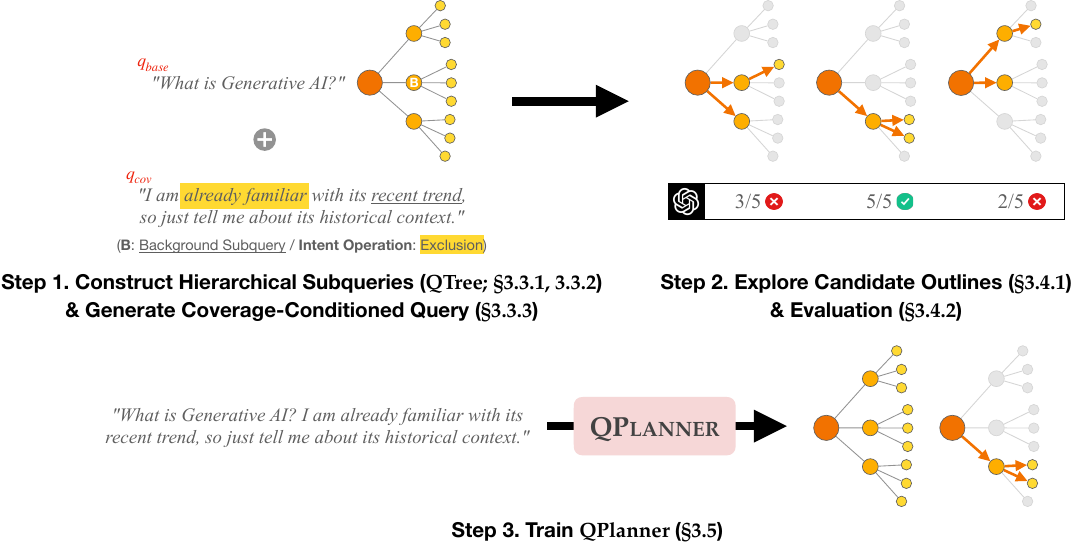}
    \caption{The overview of our framework. \texttt{\textbf{[Step 1]}} Base query ($q_{base}$) is decomposed into subqueries with diverse viewpoints (\dataset), preceded by generating coverage query ($q_{cov}$). \texttt{\textbf{[Step 2]}} After \cc candidate outlines are extracted, a judge LLM evaluates each outline and selects the best-scored one. \texttt{\textbf{[Step 3]}} Utilizing this dataset, \model is trained to sequentially generate its own \dataset and preferred outline by taking the \cc query as an input.
    }
    \label{fig:overview}
    \vspace{-3mm}
\end{figure*}

\subsection{Query Modification with LLMs}
Integrating retrieval systems with LLMs has become crucial, with query modification playing a pivotal role in improving information retrieval outcomes. Recent advancements focus on prompting LLMs to provide detailed information, such as expected documents or pseudo-answers, for query expansion~\citep{wang-etal-2023-query2doc, jagerman2023query}. Furthermore, reward signals are being used to fine-tune query modification models, optimizing search results based on the ranking of retrieved documents~\citep{ma2023query, yoon2024ask}. Additionally, complex questions are being decomposed into simpler subqueries to enhance retrieval accuracy and response generation~\citep{khot2023decomposed}.

Among various query modification strategies, query outlining stands out as an effective method for generating long responses. Outlining techniques have been primarily studied in tasks such as story generation~\citep{fan-etal-2018-hierarchical}. \citet{yang-etal-2022-re3, yang-etal-2023-doc} have also emphasized the importance of outline for narrative generation, while \citet{shao2024assisting} explored the outline as tools for presenting diverse perspectives through iterative conversational processes. More recently, \citet{lee2024navigating} improved free-form writing with outline augmentation. However, despite these advances, it has seen less attention in retrieval-augmented contexts. In addition, none of these studies systematically evaluate the generated outlines in complex scenarios (\eg, \cc scenarios). Our work aims to address this gap by proposing a controlled evaluation testbed for outlines and their impact on long-form responses.

\subsection{Evaluation of Long-form Responses}

Evaluating long-form responses from LLMs presents a significant challenge due to the subjective and multifaceted nature of the task. 
Previous studies~\citep{krishna-etal-2021-hurdles, xu-etal-2022-answer, xu-etal-2023-critical} highlight the limitations of automated metrics in accurately assessing long-form texts, underscoring the need for more nuanced evaluation methods. Several approaches have emerged to tackle this issue by incorporating multi-metric evaluation frameworks~\citep{liang2022holistic, gehrmann2023repairing, shevlane2023model, ye2023flask}, as well as task-specific metrics for fact verification and summarization~\citep{min-etal-2023-factscore, krishna-etal-2023-longeval}.
Recent research has also investigated model-based evaluations where learned models are used to generate automated scores~\citep{yuan2021bartscore, liu-etal-2023-g, kim2024prometheus}. 

While most of these studies focus solely on the evaluation of long-form responses, we extend this by evaluating both the outlines and responses they guide. Drawing from insights in cognitive psychology~\citep{Kellogg1988AttentionalOA}, we argue that outlines alleviate the cognitive overload for readers, functioning as effective content drafts and providing the core structure for long-form writing.
\section{Framework}
\label{sec:framework}

\subsection{Background}
We refer to \dataset as a tree-shaped hierarchical set of subqueries (defining ``subquery'' as each node in \dataset) derived from a single user query. We set both the depth and the width of \dataset at three levels (\ie, 3+9+27=39 subqueries in each \dataset). Additionally, as illustrated in \Cref{fig:teaser}, we define \cc query as the concatenation of the user's original query (\textit{base query}; $q_{base}$) and additional coverage-constraining query (\textit{coverage query}; $q_{cov}$), represented as $C^2 = [q_{base};q_{cov}]$.

\begin{table*}[t]
\centering
\renewcommand{\arraystretch}{0.7}
\setlength{\tabcolsep}{14pt} 
\begin{tabular*}{0.8\textwidth}{cccc} 
\toprule
\textbf{Dataset}  & \textbf{Source} &  \textbf{Train} &  \textbf{Test} \\ \midrule[1pt]
   ASQA~\citep{stelmakh-etal-2022-asqa} & Wikipedia &  4,353 &   100  \\ \midrule
   Longform~\citep{koksal2024longform} & Wikipedia &  4,483 &   100\\ \midrule
   ExpertQA~\citep{malaviya2023expertqa} & Expert &   1,741 &   100\\ \midrule[1pt]
   Total & -  & 10,577 &   300\\
\bottomrule
\end{tabular*}
\caption
{
Basic statistics of our seed datasets. We specify the number of questions in each split. We obtain $q_{base}$ from these datasets constructed from various corpus.
}
\label{tab:dataset}
\vspace{-3mm}
\end{table*}

\subsection{Overview}
\label{sec:data}

\Cref{fig:overview} illustrates the procedural framework, including the construction of  \dataset and \model. Followed by collecting $q_{base}$ (\Cref{sec:collection}), we construct \dataset (\Cref{sec:qtree_decomp}) and generate $q_{cov}$ (\Cref{sec:qtree_inst}). Generated \cc queries (\ie, $q_{base}$ and $q_{cov}$) are then utilized to select candidate outlines. For example, the answer to a $q_{base}$ \texttt{"What is Generative AI?"} can contain diverse perspectives, including its latest trend, historical context, and application across different fields. Within available outlines that guide to satisfying answers, our goal is to obtain the outline that follows $q_{cov}$ \texttt{"Tell me about its historical context"}. Therefore, within the range of \dataset, we parse candidate outlines for each \cc query (\Cref{sec:qtree_parsing}), preceded by the evaluation for selecting the optimal outline (\Cref{sec:qtree_eval}). The following subsections detail the procedural generation, and all used prompts are provided in \Cref{sec:prompts}.

\subsection{Preparing \cc Queries (\texttt{Step 1})}

\subsubsection{Base Query ($q_{base}$) Collection}
\label{sec:collection}

We first collect $q_{base}$ that requires long-form content composition to respond. Specifically, we employ two Wikipedia-based long-form question answering datasets -- ASQA~\citep{stelmakh-etal-2022-asqa} and Longform~\citep{koksal2024longform}, and one from expert domains -- ExpertQA~\citep{malaviya2023expertqa}, as demonstrated in \Cref{tab:dataset}. For the test set, we sample 100 test queries for each dataset. By leveraging LLMs\footnote{We use \texttt{gpt-4-0125-preview} of OpenAI~\citep{openai2023gpt4} with a temperature of 1.0, throughout this work.}, we construct \cc queries by combining these 10K $q_{base}$ with corresponding $q_{cov}$ (see \Cref{sec:qtree_inst}). We slightly modify and filter a few $q_{base}$ containing noises, described in \Cref{sec:seed}.

\subsubsection{\dataset Construction}
\label{sec:qtree_decomp}

Prior to generating $q_{cov}$, we decompose $q_{base}$ into diverse subqueries as a tree structure (\ie, \dataset). The purpose of constructing \dataset for each $q_{base}$ is to unfold the scope of information within parametric knowledge of LLMs. This structured graph also enables effective instruction generation (will be detailed in the next subsection) according to the hierarchy of abstractiveness. Subqueries in deeper depth present more specific subtopics. \Cref{tab:ex_fulltree} in \Cref{sec:ex_tree} illustrates an example of \dataset. 

\paragraph{Quality Check} In the query decomposition stage, we ensure that \dataset contains a predefined number of subqueries (\ie, three) in each depth and width and does not overlap each other. This can be simply done by heuristically inspecting and comparing the structured output. 

\subsubsection{Coverage Query ($q_{cov}$) Generation}
\label{sec:qtree_inst}

\begin{table*}[t]
\centering 
\renewcommand{\arraystretch}{0.4}
\setlength{\tabcolsep}{11pt}
\begin{tabular*}{\textwidth}{cl} 
\toprule
\textbf{Intent Operation} & \textbf{$q_{cov}$ Examples} \\ \midrule[1pt]
\multirow{15}{*}{\normalsize \incl} & $\blacktriangleright$ Considering my eagerness to learn about educational analysis, include \\
& \; any thematic discussions by experts on the qualifications or\\
& \; contributions of the newly appointed UPSC member to the commission. \\ \\ 
& $\blacktriangleright$ Since I'm curious about the roots of the name, please explain where\\
& \; the name Jibril originated from. \\ \\
& $\blacktriangleright$ Given my interest in agriculture, include details about how different \\
& \; seasons can enhance or diminish the quality and quantity of tea \\ 
& \; produced in various regions.
 \\ \\
\hline \\ 
\multirow{14}{*}{\normalsize \excl} & $\blacktriangleright$ Ensure you omit any irrelevant details about Mary Poppins itself; I'm \\
& \; only interested in the birth date of the actress who played the bird lady.\\ \\
& $\blacktriangleright$ Since I already understand the elements required to prove theft, ensure\\
&\; to focus on the different classifications of theft in various legal systems\\
&\; without delving into the proof elements. \\ \\
& $\blacktriangleright$ Avoid diving into the biographies of other directors from the series; \\
& \; I'm only interested in the one who directed the initial movie. \\
\bottomrule
\end{tabular*}
\caption
{
Example of generated $q_{cov}$ according to intent operations (randomly sampled from the training set).
}
\label{tab:ex_inst}
\end{table*}
\begin{table*}[t]
\centering
\renewcommand{\arraystretch}{0.9}
\begin{tabular}{l}
\toprule
\multirow{3}{*}{\shortstack[l]{\textbf{\cc Query}: \\ Describe the film The Woman Hunt. Since I'm already familiar with how audiences and critics\\ received The Woman Hunt, please avoid discussing reviews or reception in your explanation.}} \\ \\ \\ 
\midrule[1pt]
\begin{minipage}{\textwidth}
{\textbf{Parsed Outline:}}
{\begin{lstlisting}[basicstyle=\normalsize]
1. What is the plot of The Woman Hunt?
    1.1. What are the main events in The Woman Hunt?
        1.1.1. What initiates the conflict in The Woman Hunt?
        1.1.2. What is the climax of The Woman Hunt?
\end{lstlisting}}
\end{minipage} \\
\bottomrule
\end{tabular}
\caption{
Example of parsed outline. Example of corresponding \dataset is available at \Cref{tab:ex_fulltree}.
}
\label{tab:ex_subtree}
\vspace{-3mm}
\end{table*}

To remind, the role of $q_{cov}$ is to specify certain subtopics to address (\ie, include or exclude) within a broad range of information. Therefore, generating $q_{cov}$ from \dataset requires selecting a specific viewpoint to cover. However, solely relying on LLMs' parametric knowledge does not guarantee the diversity of realistic situations. We therefore adopt the following two concepts to assist in generating $q_{cov}$.

\begin{itemize}
    \item \textbf{\textit{Background Subquery}}: Understandably, asking for specific knowledge means that users are recognizing the knowledge itself. With this in consideration, we randomly select a single subquery from \dataset, which will be the knowledge users are aware of. We define this subquery as \textbf{background subquery}. The specificity of the background subquery differs according to the depth of the selected query. 
    
    \item \textbf{\textit{Intent Operation}}: While considering a particular subject to ask, users may choose whether the content should be addressed within the responses. We conceptualize user intent through a binary operation (\ie, \incl, \excl), thereby facilitating the generation of $q_{cov}$ that explicitly request the inclusion/exclusion of the subtopic on the background subquery.
\end{itemize}

In practice, we prompt LLM to generate $q_{cov}$ by combining a randomly selected background subquery from \dataset with intent operation\footnote{Although we use background subquery to generate $q_{cov}$ in this section, it is also used to construct baselines. Refer to \Cref{sec:baselines}.}. As demonstrated in \Cref{tab:ex_inst}, combinations of background subquery and intent operation yield diverse $q_{cov}$ for each $q_{base}$. Especially, requirements of $q_{cov}$ with \excl operation are more complicated (\eg, avoiding one topic but focusing on another topic) than \incl. We analyze the performance difference according to intent operations in \Cref{sec:analysis_intent}. We sample five preliminary $q_{cov}$ per each $q_{base}$ and finally choose one if corresponding three candidate outlines are parsed correctly (which will be further described in Section \ref{sec:qtree_parsing}). 

\subsection{Exploring Candidate Outlines \& Evaluation  (\texttt{Step 2})}

\subsubsection{Parsing Outlines}
\label{sec:qtree_parsing}

In this stage, LLM sequentially extracts JSON-formatted candidate outlines from \dataset that satisfy instructions of \cc queries\footnote{Our preliminary verification identifies that sequentially generating candidate outlines shows more diversity than temperature-based sampling. Refer to \Cref{sec:sampling} for case studies.}. \Cref{tab:ex_subtree} visualizes the example of a candidate outline, consisting of hierarchical subqueries (\ie, plot - main event - conflict \& climax) about \textit{The Woman Hunt}. We extract three candidate outlines per each \cc query. 

\paragraph{Quality Check} We fix the number of subqueries within each outline to four, guaranteeing that all subqueries are directly connected or neighboring within \dataset, as illustrated in \Cref{tab:ex_subtree}. Additionally, we verify the JSON parsability of each outline and ensure that all subqueries do not overlap each other. For the efficient usage of API calls, we heuristically remove subqueries in leaf nodes if an outline contains more than four subqueries and include the outline as a candidate.

\subsubsection{Evaluating Outline Quality}
\label{sec:qtree_eval}

In order to rank three candidate outlines, we leverage LLM (\texttt{gpt-4-0125-preview}) to serve as a judge deciding whether the content on each candidate outline follows \cc query. More precisely, we prompt the model to assign five-point Likert-scale scores with rationales, measuring how faithfully the outline aligns with the \cc query. Since outlines are significantly shorter than long-form text while maintaining core contents~\citep{Kellogg1988AttentionalOA}, it is expected that evaluating outlines is more efficient and intuitive than directly evaluating long responses. These scored outlines are utilized as supervision and alignment pairs for training \model, which will be described in further sessions.

\subsection{Training \model (\texttt{Step 3})}
\label{sec:qplanner}

To generalize with arbitrary \cc queries, we train a 7B language model named \model. We instruct \model to sequentially generate \dataset and select an outline, as we intend that \dataset serves like an intermediate Chain-of-Thought~\citep{NEURIPS2022_9d560961} reasoning process. More technical details are described in \Cref{sec:qplanner_detail}.

\section{Experiments}


\begin{table}[t]
\centering 
\renewcommand{\arraystretch}{0.8} 
\setlength{\tabcolsep}{7pt} 
\begin{tabular*}{\columnwidth}{lcc} 
\toprule
& \textbf{Mean} ($\uparrow$) & \textbf{SD} ($\downarrow$) \\ \midrule[1pt]
\textbf{Random Basline} & 2.57 & 1.44 \\ \midrule
\textbf{SFT-\model} & \multirow{2}{*}{2.79} & \multirow{2}{*}{1.40} \\
(31K) & & \\ \midrule
\textbf{DPO-\textsc{SynNeg}} & \multirow{2}{*}{2.98} & \multirow{2}{*}{1.39} \\
(31K + 8K align) & & \\ \midrule
\textbf{DPO-\textsc{Combined}} & \multirow{2}{*}{3.01} & \multirow{2}{*}{1.36} \\
(31K + 16K align) & & \\ \midrule
\textbf{\textsc{DPO}-\model} & \multirow{2}{*}{\textbf{3.16}} & \multirow{2}{*}{\textbf{1.33}} \\
(Ours; 31K + 8K align) & & \\
\bottomrule
\end{tabular*}
\caption
{
Mean and standard deviation (SD) for automatic outline evaluation (five-point Likert scale). DPO-\model scores the highest mean score and the lowest SD, indicating robust improvement.
}
\label{tab:eval_automatic}
\end{table}

\subsection{Training Details of \model}
\label{sec:qplanner_detail}
We employ supervised fine-tuning (SFT) and alignment tuning for the training \model. First, we train the Llama-2-7B-Chat model~\citep{touvron2023llama} using 10K \cc queries mapped with 31K candidate outline pairs, constructed through \Cref{sec:qtree_parsing}. This training phase allows the model to generate formatted outlines following \cc queries (named \textbf{SFT-\model} hereafter). 

Then we further align the preferred outline by adopting a variant of direct preference optimization (DPO)~\citep{rafailov2023direct}. Following \citet{tunstall2023zephyr} that show the possibility of distilling the preference of large teacher models into a targeted model, we utilize LLM evaluation scores previously acquired in \Cref{sec:qtree_eval} as reward signals for aligning \model (named \textbf{DPO-\model} hereafter). We regard the highest-scored outline as a positive (chosen) sample and the lowest-scored outline as a negative (rejected) sample. We skip samples whose highest and lowest scores are the same in the alignment stage. 

The amount of the final training sample is 31,488 for SFT-\model and 8,568 for DPO-\model, respectively. Refer to \Cref{sec:training} for further details, such as hyperparameters.

\subsection{Baselines for Outline Comparison}
\label{sec:baselines}
\paragraph{Random Baseline} Since the output of \model accompanies \dataset as an intermediate reasoning process, we can extract an arbitrary outline by leveraging this, regardless of the \cc queries. Specifically, we select a random background subquery from \dataset generated by SFT-\model, then extend the branch to randomized directions (\ie, upper depth, neighbor, or lower depth) until four subqueries are connected as a single outline. Intent operation is not considered in this random baseline. 

\paragraph{DPO-\textsc{SynNeg}} To further explore the effectiveness of selected (\ie, LLM-scored) negative samples in DPO-\model, we prepare another DPO model trained with different types of negative samples. While negative samples of DPO-\model are based on LLM scores, we can also heuristically synthesize negative samples with \dataset, background subquery, and intent operation. This procedure is similar to generating random baseline, except for ensuring that synthesized outlines have the \textit{opposite} intent operation to the original intent. For example, if the positive outline \textit{includes} background subquery, the synthesized outline is designed to \textit{exclude} that subquery by selecting another random background subquery within \dataset. For the opposite situation, the synthesized outline must contain background subquery of the positive outline. On the 8K DPO-\model training set, we maintain the positive samples and replace negative samples with synthetically generated outlines.

\paragraph{DPO-\textsc{Combined}} We also measure the performance of combining negative samples in DPO-\model and DPO-\textsc{SynNeg}. That is, the number of training samples is doubled.

\section{Results}
\label{sec:results}

\subsection{Automatic Outline Evaluation}

We prompt LLM (\texttt{gpt-4-0125-preview)} to score generated outlines in the test set. We use scoring rubric in \Cref{tab:rubric} in \Cref{sec:outline_eval}.

\subsubsection{Mean Score Comparison}

\Cref{tab:eval_automatic} shows our test result with a five-point Likert scale. We score outlines generated by each trained model\footnote{A few cases return an outline with 3 or 5 queries, which is not an ideal number of the output (\ie, 4), but we do not filter them in our evaluation.}, focusing on whether the content of outlines follows given \cc queries (as mentioned in \Cref{sec:qtree_eval}). 

We find that the random baseline shows the lowest mean score (2.57) with the highest standard deviation (1.44) on our test set. While SFT-\model shows a higher score than the random baseline, we find that DPO-\model significantly improves the score (3.16) with the lowest standard deviation (1.33). It implies that we can \textbf{leverage LLM-generated scores as reward signals in query outlining} when constructing positive-negative pairs, even in the absence of explicit and gold reward criteria for their construction~\citep{ma2023query, yoon2024ask}.

On the comparison with DPO-\model and DPO-\textsc{SynNeg}, we observe that \textbf{negative samples of DPO-\model are notably more effective} than the other one, since the only difference between them is the type of negative samples. We conjecture that constructing ``hard'' negatives\footnote{We define LLM-generated negative samples as ``hard'' when compared to synthetic negative samples whose intent operations are explicitly opposite to positive outlines.} (\ie, less scored subtrees with the ``same'' intent) is an important factor to align with \cc queries, sharing insights with different studies on hard negatives~\citep{rosset-etal-2023-axiomatic, scarlatos2024improving}. Regarding DPO-\textsc{Combined}, the performance becomes worse than DPO-\model despite the doubled amount of alignment pairs, implying the importance of selective negative samples.

\begin{figure}[t]
    \centering
    \includegraphics[width=\columnwidth]{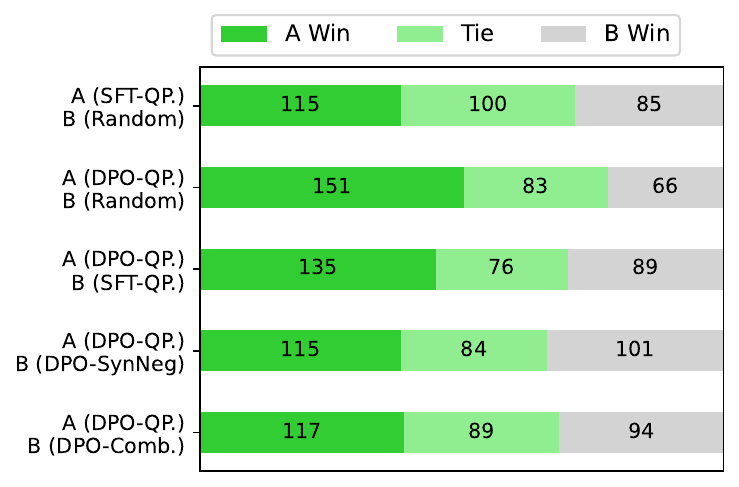}
    \caption{Pairwise comparison for each \cc query in automatic outline evaluation.}
    \label{fig:win}
\end{figure}

\subsubsection{Pairwise Comparison}

For comprehensive evaluation, we also compare pairwise scores among models. As illustrated in \Cref{fig:win}, \textbf{generated outlines of DPO-\model are more preferred than all other baselines on the same \cc query}, which is aligned with the atomic scoring result in \Cref{tab:eval_automatic}. Given the fact that exhaustively devising preferred outlines for \cc queries is labor-intensive, our \model is an effective solution for exploring and creating preferred outlines for long-form responses.

\subsection{Human Outline Evaluation}

We conduct a human study to identify the effectiveness of \model. We describe detailed experimental setup in \Cref{sec:humanstudy}, such as demographics and payment.

We let participants read and rate two outlines generated from SFT- and DPO-\model for randomly selected 100 \cc queries from the test set using the same five-point Likert scale criteria as the automatic evaluation. Each evaluator rates from 4 to 10 outlines in random order (to avoid position bias), and each outline has at least 6 evaluators (Average: 6.55, Max: 15). We intend to have as many evaluators as possible to rate individual outlines to gather a collective rating for each set. This is because even with simple outlines, judging outlines with unfamiliar topics is a highly intellectual and unavoidably subjective task.

Consequently, we find \textbf{significant positive correlations between human-rated scores and LLM-rated scores} -- both for SFT (Pearson's $r$ = 0.51, p-value $<$ 0.001) and DPO-\model (Pearson's $r$ = 0.39, p-value $<$ 0.001), which indicates positive relationships with large and medium strength, respectively. Moreover, DPO-\model receives higher human scores (Mean= 3.29, Std=0.81) than SFT (Mean= 3.03, Std=0.78)\footnote{This trend is supported even when we regress scores on model version (SFT or DPO-\model) and the total length of outlines in characters, with outline id as a fixed effect (Model:$b$=0.27, p-value=0.01; Outline length: $b$=0.01, p-value=0.30). That is, the length of outlines is not predictive of scores.}. We report that evaluating highly subjective tasks may introduce varied ratings among human evaluators despite assigning a large number of evaluators to derive the majority opinion (Krippendorff's $\alpha$: SFT-\model = 0.22; DPO-\model = 0.23), as observed in other studies~\citep{rottger-etal-2022-two, abercrombie2023consistency}.

\subsection{Human RAG Evaluation}

For long-form response evaluation, we do not automatically measure due to the lack of reliability in long-form text evaluation~\citep{xu-etal-2023-critical}. Instead, we recruit another participant to validate the effectiveness of \model on RAG downstream tasks\footnote{Details such as recruitment, instructions, and compensation are described in Appendix \ref{sec:ragevaldetail}}. 100 RAG responses from the test set are sampled for evaluation, and ten evaluators are assigned for each response. Following insights from~\citet{kim2023understanding} where the writing format of model responses affects human preferences, we fix the response format with Markdown to compare responses by focusing only on their content. In addition, we prompt LLM\footnote{\texttt{gpt-4-0125-preview} is used.} to generate responses by strictly relying on given evidence to prevent LLM from arbitrarily responding with its parametric knowledge. We assume web search scenarios for the RAG setup, providing detailed information in \Cref{sec:retrieval}.

\begin{figure}[t!]
    \centering
    \begin{subfigure}[b]{\columnwidth}
        \centering
        \includegraphics[width=\textwidth]{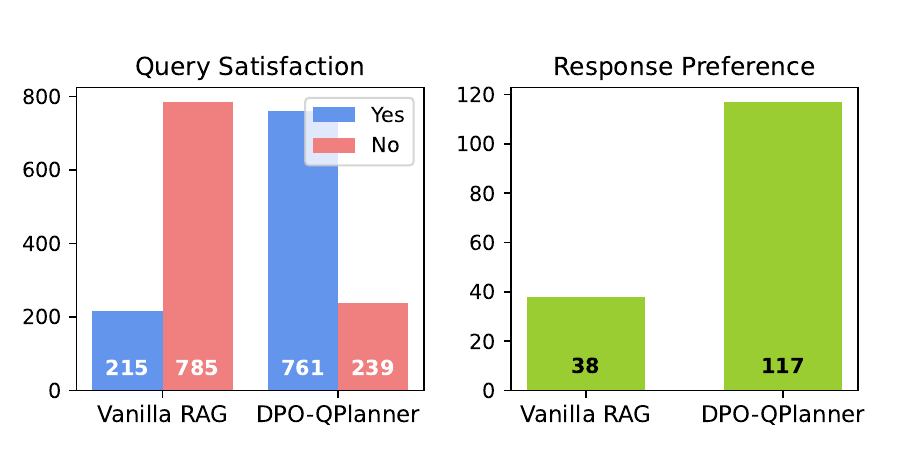}
        \caption{\model as Search Query}
        \label{fig:study_search}
    \end{subfigure}
    \hfill 
    \begin{subfigure}[b]{\columnwidth}
        \centering
        \includegraphics[width=\textwidth]{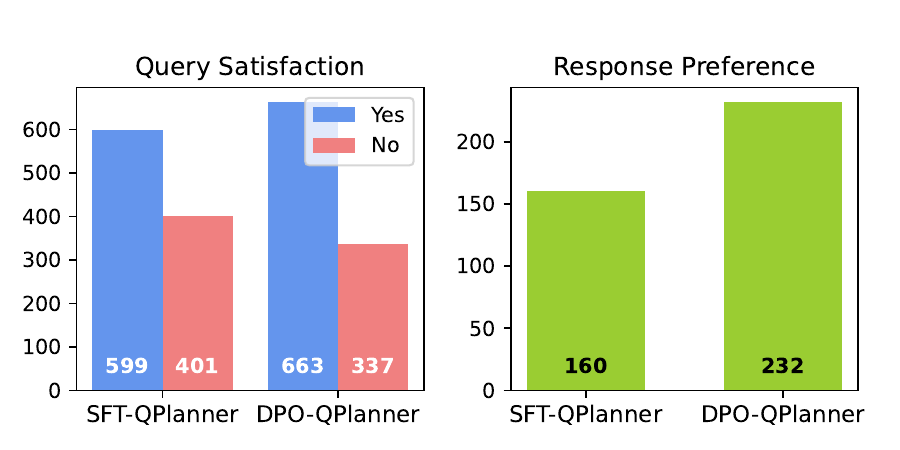}
        \caption{\model as Content Draft}
        \label{fig:study_outline}
    \end{subfigure}
    \caption{Human evaluation results.}
    \label{fig:humanstudy}
\end{figure}

Regarding evaluation criteria, we first instruct participants to judge whether generated responses follow requirements of \cc queries or not (\texttt{Query Satisfaction} in \Cref{fig:humanstudy}). We guide them to annotate "Yes" if responses at least partially address topics within \cc queries. For response pairs annotated as "Yes" in both models, participants select their preferred response (\texttt{Response Preference} in \Cref{fig:humanstudy}).

\paragraph{\model as Better Search Query}
\label{sec:resp_eval_search}

We verify whether subqueries within outlines can help search relevant documents. We compare responses of vanilla RAG with those of DPO-\model using the exactly same prompt. That is, subqueries of DPO-\model only affect the search result. As illustrated in \Cref{fig:study_search}, we observe that the conventional RAG pipeline does not properly retrieve relevant evidence for answering \cc queries, whose requirements are far more complicated than normal queries. Furthermore, among responses that satisfy the requirements of \cc queries, responses of DPO-\model are mostly preferred.

\paragraph{\model as Better Content Draft}
\label{sec:resp_eval_outline}

We further investigate whether better outlines lead to better responses. In this setup, we compare responses of SFT-\model and DPO-\model. The exactly same prompt is used for this comparison, and subqueries within outlines are included in the prompt for composing responses and retrieving documents. Results in \Cref{fig:study_outline} indicate that further aligning \model with preference can provide preferred outlines, while SFT-\model also shows a similar tendency with DPO-\model.

Since the outcome of both studies is binary (\ie, satisfactory or not), and each human evaluator judges both responses for the same query (within-subjects design), we conduct two McNemar's tests~\citep{mcnemar1947sampling} to examine whether the differences we find are statistically significant. The contingency tables used for the tests can be found in \Cref{tab:contingency}. The results confirm that DPO-\model significantly outperforms Vanilla RAG (test-statistics = $60$, p-value < 0.001) and SFT-\model (test-statistics = $207$, p-value = 0.004).

\section{Conclusion}

In this work, we suppose complicated user scenarios asking for a constrained range of a specific topic, called coverage-conditioned (\cc) query scenarios. To simulate \cc scenarios and a controlled environment for creating and evaluating query outlines (\textit{RQ1}), we construct \dataset, hierarchical sets of subqueries representing diverse perspectives of the original query. Playing a role as boundaries for available outlines, \dataset allows systematic comparison of diverse outlines. Subsequently, we train \model which extracts customized outlines from \dataset for \cc queries. Regarding our \textit{RQ2}, our findings based on automatic and human evaluation show that (1) preference-aligned \model can generate better outlines, (2) outlines enable improved document search, and (3) better outlines lead to preferred responses. We believe our work shows the possibility of \dataset as a testbed for exploring effective pre-writing strategies to deal with complicated queries.

\section*{Limitations}
\label{sec:limitation}

We discuss the current limitations of our work. First, our graphical representation of subquery nodes adheres to canonical tree structures, with each node connected to three child nodes, but it can be adjusted (\ie, composing more or less subqueries) according to tasks or domains. For example,  in a complex domain like medical diagnosis, a larger number of subqueries might be necessary to cover various symptoms, possible conditions, diagnostic tests, and treatment options. In contrast, for a straightforward factual query in a domain like mathematics, fewer subqueries might be sufficient to reach a comprehensive answer. Identifying this optimal number still remains an open question and represents a promising direction for future investigation. We believe that our experimental setup serves as an initial testbed for validating these research questions.

It should also be noted that the contents of retrieved documents in our RAG setup can affect the detailed factual consistency of final responses. Although we set the same search configuration among all methodologies, additional fact verification of documents and responses is still needed for practical applications.

While demonstrating significant performance gains both in automatic and human judgements, we find that state-of-the-art LLMs still have difficulty generating long-form responses that handle detailed coverage of \cc  queries. This is presumably due to the complexness of \cc queries, and it arises the importance of constructing meticulous benchmarks evaluating long-form responses to complicated queries. This will be another direction of the future work. Lastly, we would like to mention that our five-point scoring schema can be further improved by considering multiple aspects with a fine-grained score rubric.

\section*{Ethical Considerations}
\label{sec:ethical}

Since our \dataset is generated with benchmarks based on Wikipedia and domain experts, we do not filter sensitive or unsafe contents throughout our studies. For the practical application in the future, deliberate content selection will be required for the safety. In addition, we explicitly share our experimental setup of human studies for transparency in \Cref{sec:humanstudy}.

\bibliography{custom}

\clearpage

\appendix
\onecolumn

\section{Base Query Modification}
\label{sec:seed}

For the ASQA\footnote{Apache 2.0 License}~\citep{stelmakh-etal-2022-asqa} and ExpertQA\footnote{MIT License}~\citep{malaviya2023expertqa} dataset, we do not modify the base query. For the Longform\footnote{MIT License}~\citep{koksal2024longform} dataset, as there are additional format-related instructions concatenated with the base query (\eg, \textit{Respond in 3 sentences.}), we eliminate them by using regular expressions. Moreover, we find that Longform dataset contains noisy queries (\eg, \textit{This does not provide enough information for an answer to be provided.}), which are unfiltered artifacts generated by large language models. In this case, we manually filter similar expressions.

\section{Outline Sampling Comparison}
\label{sec:sampling}

To identify the effectiveness of sequentially generating candidate outlines at once, we generate candidate outlines using temperature sampling. As shown in \Cref{tab:sampling}, sequential generation meaningfully diversifies candidate outlines than temperature sampling. Similar results are obtained when we increase the sampling temperature.

\begin{table*}[t]
\centering
\renewcommand{\arraystretch}{1.0}
\begin{tabular}{l}
\toprule
\multirow{3}{*}{\shortstack[l]{\textbf{\cc Query}: \\ When was the museum of islamic art built? Since I'm already familiar with who the lead architect\\ of the Museum of Islamic Art was and their design vision, focus on providing the construction \\date without delving into architectural design specifics.}} \\ \\ \\ \\
\midrule[1pt]
\begin{minipage}{0.95\columnwidth}
\textbf{Parsed Outline (Sampling Temperature 1.0):} 
{\begin{lstlisting}
3.2. Which construction company undertook the building of the Museum of Islamic Art, and in which year did they commence work?
    3.2.1. What firm was responsible for the construction of the Museum of Islamic Art, and when did they start?
    3.2.2. Can you identify the construction firm that built the Museum of Islamic Art and the year construction began?
    3.2.3. Who constructed the Museum of Islamic Art, and in what year did construction efforts begin?

3.3. What were the financial and logistical challenges in constructing the Museum of Islamic Art?
    3.3.1. What budgetary constraints affected the construction of the Museum of Islamic Art?
    3.3.2. What logistical hurdles were encountered during the building of the Museum of Islamic Art?
    3.3.3. In constructing the Museum of Islamic Art, what were the major financial and logistical issues

3. Who contributed to the design and construction of the Museum of Islamic Art?
    3.2. Which construction company undertook the building of the Museum of Islamic Art, and in which year did they commence work?
    3.2.1. What firm was responsible for the construction of the Museum of Islamic Art, and when did they start?
    3.2.2. Can you identify the construction firm that built the Museum of Islamic Art and the year construction began?
 \end{lstlisting}}
\end{minipage} \\ \midrule

\begin{minipage}{0.95\columnwidth}
\textbf{Candidate Outlines (Sequential Generation):} 
{\begin{lstlisting}
2. When did the Museum of Islamic Art open to the public?
    2.1. What is the official opening date of the Museum of Islamic Art?
        2.1.1. On which date was the Museum of Islamic Art inaugurated?
        2.1.2. When was the grand opening of the Museum of Islamic Art?

1. What is the construction date of the Museum of Islamic Art?
    1.2. Who was the architect behind the Museum of Islamic Art design, and in what year did the construction take place under their guidance?
        1.2.1. What is the name of the architect who designed the Museum of Islamic Art, and during which year was their design realized?
        1.2.3. In what year did construction of the Museum of Islamic Art occur under the designated architect's design?

3. Who contributed to the design and construction of the Museum of Islamic Art?
    3.2. Which construction company undertook the building of the Museum of Islamic Art, and in which year did they commence work?
        3.2.1. What firm was responsible for the construction of the Museum of Islamic Art, and when did they start?
        3.2.3. Who constructed the Museum of Islamic Art, and in what year did construction efforts begin?

 \end{lstlisting}}
\end{minipage} \\

\bottomrule
\end{tabular}
\caption{
Comparison of temperature sampling and sequential generation of candidate outlines.
}
\label{tab:sampling}
\end{table*}

\clearpage

\section{Example of \dataset}
\label{sec:ex_tree}
\begin{table*}[ht]
\centering
\renewcommand{\arraystretch}{1.0}
\begin{tabular}{l}
\toprule
\multirow{3}{*}{\shortstack[l]{\textbf{\cc Query}: \\ Describe the film The Woman Hunt. Since I'm already familiar with how audiences and critics\\received The Woman Hunt, please avoid discussing reviews or reception in your explanation.}} \\ \\ \\ 
\midrule[1pt]
\begin{minipage}{0.95\columnwidth}
\textbf{\dataset:} 
{
\begin{lstlisting}
1. What is the plot of The Woman Hunt?
    1.1. What are the main events in The Woman Hunt?
        1.1.1. What initiates the conflict in The Woman Hunt?
        1.1.2. What is the climax of The Woman Hunt?
        1.1.3. How does The Woman Hunt end?
    1.2. Who are the main characters in The Woman Hunt?
        1.2.1. Who is the protagonist of The Woman Hunt?
        1.2.2. Who is the antagonist in The Woman Hunt?
        1.2.3. What supporting characters play crucial roles in The Woman Hunt?
    1.3. What themes are explored in The Woman Hunt?
        1.3.1. What is the primary theme of The Woman Hunt?
        1.3.2. How does The Woman Hunt explore gender dynamics?
        1.3.3. What messages does The Woman Hunt convey about survival?
2. Who directed The Woman Hunt?
    2.1. What is the directorial style of The Woman Hunt?
        2.1.1. How does the director use camera angles in The Woman Hunt?
        2.1.2. What unique directorial choices are made in The Woman Hunt?
        2.1.3. How does the pace affect the narrative in The Woman Hunt?
    2.2. What other films has the director of The Woman Hunt made?
        2.2.1. What are the most popular films by The Woman Hunt's director?
        2.2.2. How do other films by the director compare to The Woman Hunt?
        2.2.3. What recurring themes appear in the director's filmography?
    2.3. How has the director's background influenced The Woman Hunt?
        2.3.1. What aspects of the director's personal life reflect in The Woman Hunt?
        2.3.2. How does the director's cultural background inform The Woman Hunt?
        2.3.3. What previous experiences of the director shaped The Woman Hunt?
3. How was The Woman Hunt received by audiences and critics?
    3.1. What are the critical reviews of The Woman Hunt?
        3.1.1. How do film critics analyze The Woman Hunt?
        3.1.2. What are the predominant critiques of The Woman Hunt?
        3.1.3. Are there any notable defenses of The Woman Hunt's thematic choices?
    3.2. What is the audience's reaction to The Woman Hunt?
        3.2.1. How do audience perspectives on The Woman Hunt vary?
        3.2.2. What aspects of The Woman Hunt resonate most with audiences?
        3.2.3. What fan opinions of The Woman Hunt diverge from critical reviews?
    3.3. Has The Woman Hunt won any awards or recognition?
        3.3.1. What awards or nominations has The Woman Hunt received?
        3.3.2. How does The Woman Hunt rank among other films of its genre?
        3.3.3. Are there any film festivals where The Woman Hunt was highlighted?
 \end{lstlisting}}
\end{minipage} \\
\bottomrule
\end{tabular}
\caption{
Example of \dataset generated by the process described in \Cref{sec:qtree_decomp}.
}
\label{tab:ex_fulltree}
\end{table*}

\clearpage

\section{Analysis on Intent Operations}
\label{sec:analysis_intent}
\begin{table*}[ht]
\centering
\renewcommand{\arraystretch}{0.8}
\setlength{\tabcolsep}{20pt}
\begin{tabular*}{\textwidth}{cccccc} 
\toprule
  & \multicolumn{2}{c}{\textbf{SFT-\model}} & \multicolumn{2}{c}{\textbf{DPO-\model}}  \\
  & \incl & \excl & \incl & \excl \\  \midrule[1pt]
   Mean & 2.85 &  2.74 & 3.22 & 3.10  \\ \midrule
   SD & 1.23 &  1.55  & 1.15 & 1.47 \\ 
\bottomrule
\end{tabular*}
\caption
{
Mean and standard deviation (SD) according to the intent operation in automatic outline evaluation.
}
\label{tab:intent}
\end{table*}

We decompose the result of \Cref{tab:eval_automatic} according to intent operations in \Cref{tab:intent}, focusing on SFT-\model and DPO-\model scores. We discover that \cc queries based on \excl score lower than those on the intent of \incl. This result aligns with our assumption in \Cref{sec:qtree_inst}, where $q_{cov}$ with \excl operation require more complicated selection of desirable outline.

\section{Used Prompts}
\label{sec:prompts}

We use the following prompts in our work.
\begin{figure*}[ht]
    \centering
    \includegraphics[width=0.9\textwidth]{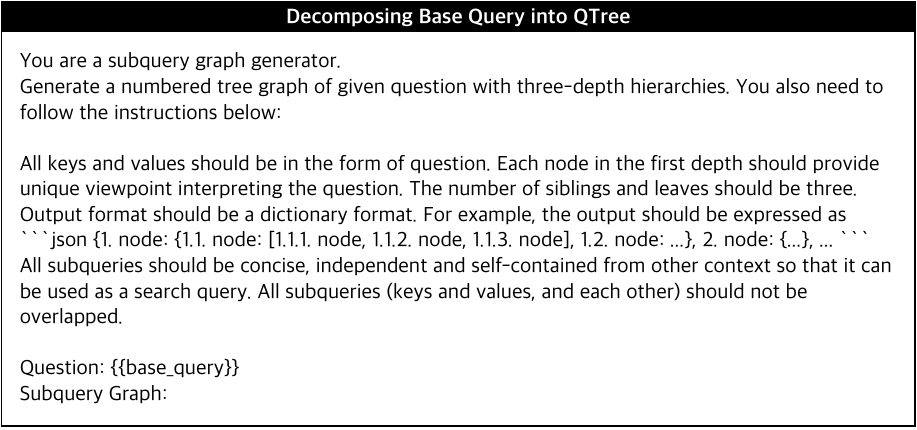}
    \vspace{-2mm}
\end{figure*}

\begin{figure*}[ht]
    \centering
    \includegraphics[width=0.9\textwidth]{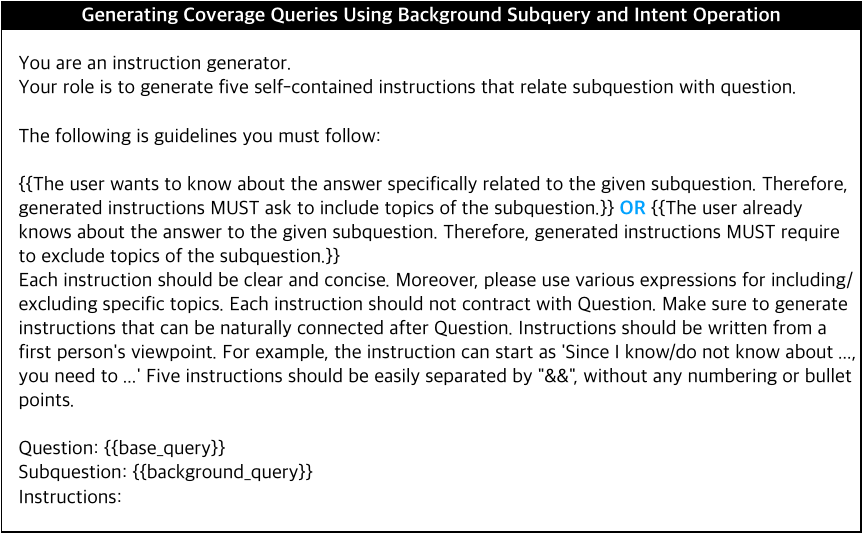}
    \vspace{-2mm}
\end{figure*}

\begin{figure*}[ht]
    \centering
    \includegraphics[width=0.9\textwidth]{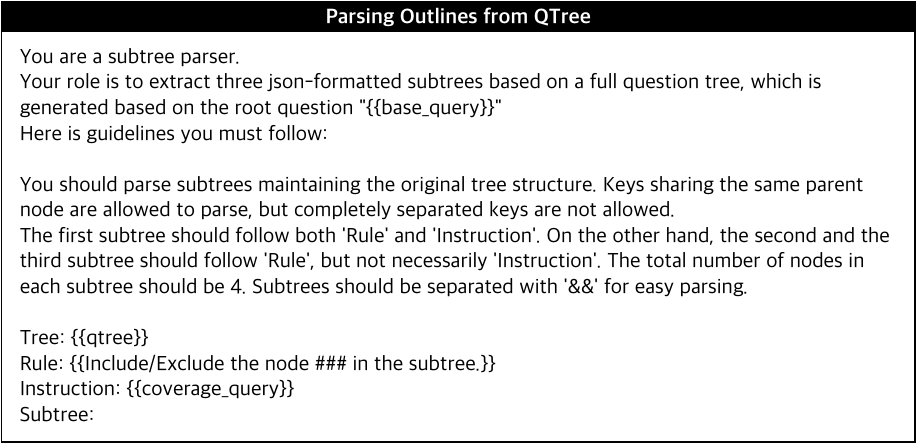}
    \vspace{-2mm}
\end{figure*}

\begin{figure*}[ht]
    \centering
    \includegraphics[width=0.9\textwidth]{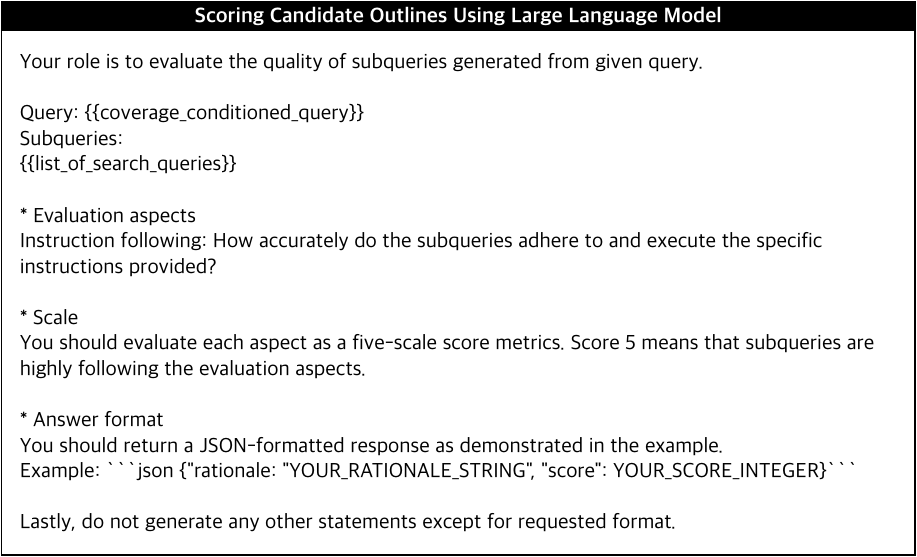}
    \vspace{-2mm}
\end{figure*}

\begin{figure*}[ht]
    \centering
    \includegraphics[width=0.9\textwidth]{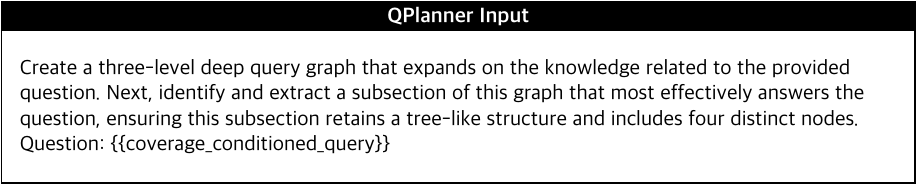}
    \vspace{-2mm}
\end{figure*}

\begin{figure*}[ht]
    \centering
    \includegraphics[width=0.9\textwidth]{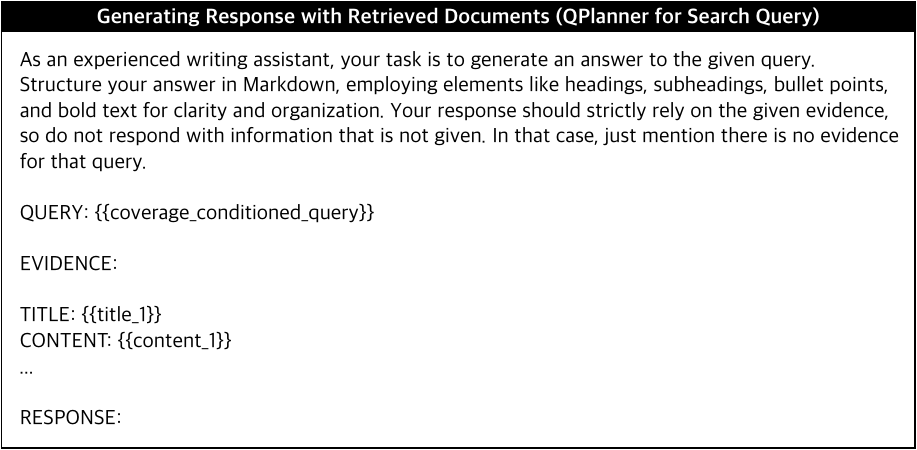}
    \vspace{-2mm}
\end{figure*}

\begin{figure*}[ht]
    \centering
    \includegraphics[width=0.9\textwidth]{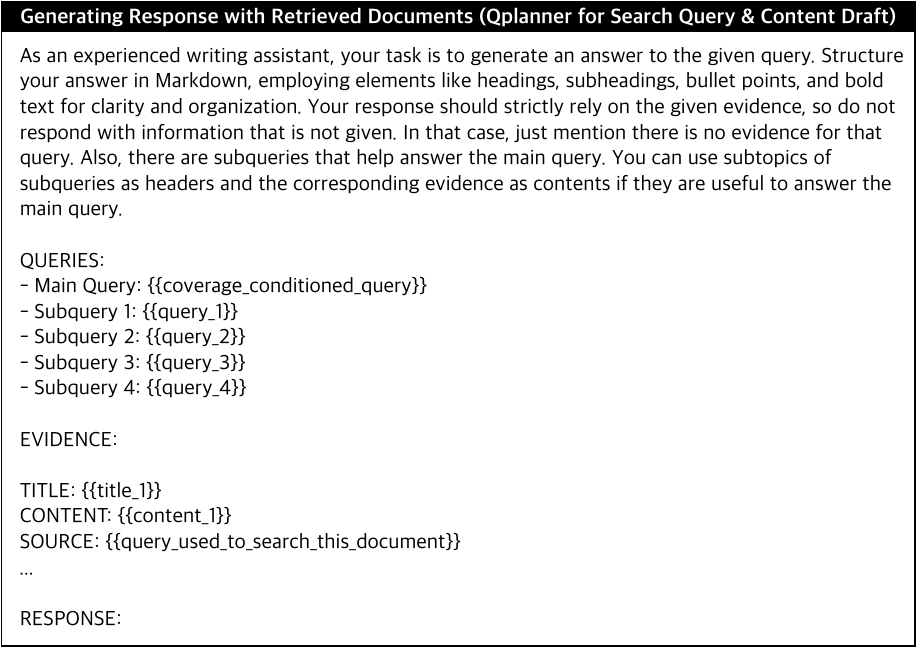}
    \vspace{-2mm}
\end{figure*}

\section{Training Details}
\label{sec:training}
\begin{table}[ht]
\centering 
\renewcommand{\arraystretch}{0.8}
\setlength{\tabcolsep}{20pt}
\begin{tabular*}{0.7\columnwidth}{ccc} 
\toprule
\textbf{Hyperparameter}  &  \textbf{SFT} &  \textbf{DPO} \\ \midrule[1pt]
   Epoch &   1 &   1  \\ \midrule
   Batch Size Per Device &  14 &   8 \\ \midrule
   Learning Rate (LR) &  2e-5 &   5e-7 \\ \midrule
   LR Schedule &  Cosine &   Cosine \\ \midrule
   Warmup Ratio &  0.1 &   0.1 \\ \midrule
   Gradient Accumulation Step &  1 &   2 \\ \midrule
   Beta &  - &   0.01 \\
   \midrule[1pt]
   \# of Samples &  31,488 &  8,568 \\ 
\bottomrule
\end{tabular*}
\caption
{
Hyperparameters for training \model. A few noisy samples are filtered in advance at SFT stage.
}
\label{tab:hparam}
\end{table}

We utilize publicly available software\footnote{\url{https://github.com/huggingface/alignment-handbook}} in our finetuning and alignment stage. We train each stage one epoch using 16 NVIDIA A100 GPUs (40GB of memory). \Cref{tab:hparam} indicates detailed hyperparameters for each stage. 

\section{Additional Information on Human Evaluation}
\label{sec:humanstudy}

\begin{figure*}[ht]
    \centering
    \includegraphics[width=0.7\textwidth]{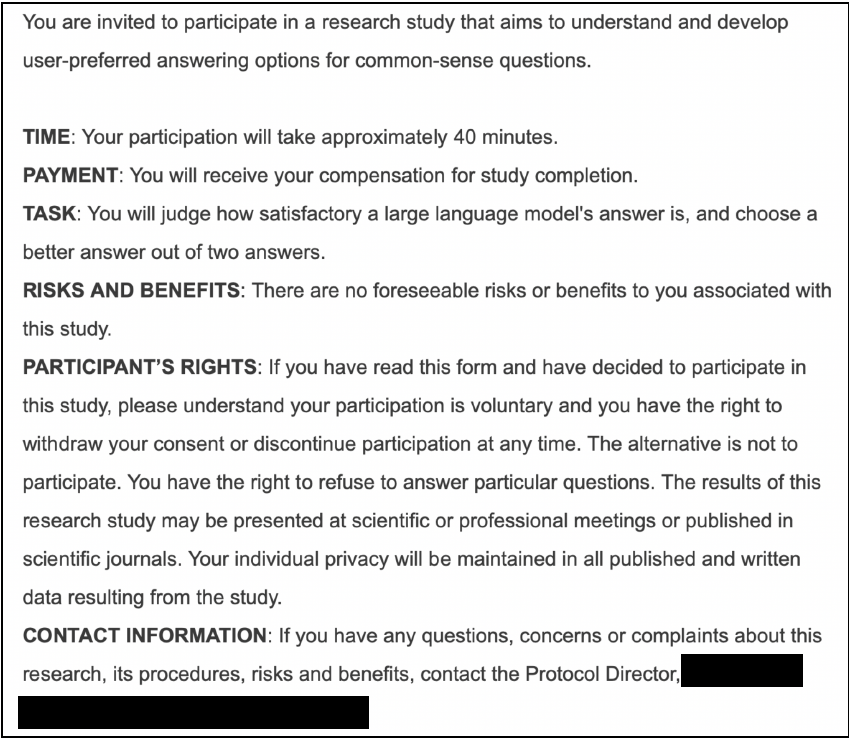}
    \caption{Initial information provided to participants in our human study.}
    \label{fig:guideline}
\end{figure*}

We conduct two human evaluation studies separately (\ie, outline and RAG evaluation). For both studies, crowdworkers are recruited from Prolific\footnote{\url{https://www.prolific.com/}}. At the beginning of the evaluation, workers are informed what task they are expected to do, there are no foreseeable benefits and risks, their participation is voluntary, and they can leave if they want (see \Cref{fig:guideline}).

\subsection{Outline Evaluation}
\label{sec:outline_eval}
\begin{table*}[ht]
\centering
\renewcommand{\arraystretch}{1.0}
\begin{tabular}{l}
\toprule
\textbf{Score Rubric} \\
\midrule[1pt]
\begin{minipage}{0.9\columnwidth}
{\begin{lstlisting}
1: The sub-questions/responses entirely disregards the instructions, providing content unrelated to the instruction.

2: The sub-questions/responses show a superficial attempt to follow instructions but significantly strays from the intended task, missing key objectives.

3: The sub-questions/responses generally adheres to the instructions but overlooks certain details or nuances, achieving only a partial match with the instruction.

4: The sub-questions/responses is closely aligned with the instructions, exhibiting minor deviations that slightly affect the completeness of the execution.

5: The sub-questions/responses exhibits impeccable adherence to the instructions, capturing all nuances and completing the task as specified.
 \end{lstlisting}}
\end{minipage} \\
\bottomrule
\end{tabular}
\caption{
Score rubric for evaluating subqueries and responses in human evaluation.
}
\label{tab:rubric}
\end{table*}

A total of 127 crowd workers participate in the evaluation (Gender: 68 men, 57 women, and 2 non-binary; Age: Mean=28.6 yrs, SD=7.9 yrs, Min=18 yrs, Max=63 yrs; Ethnicity: White: 69, Black: 44, Mixed: 11, and Asian:2; Country of residence: South Africa: 52 (41.27\%), Portugal: 20 (15.87\%), Poland: 10 (7.94\%), United Kingdom: 5 (3.97\%), Mexico: 5 (3.97\%), and 19 other countries; Highest education level completed: A majority of the evaluators hold at least a Bachelor's degree (n=83, 65.87\%)). Individual crowd workers evaluate different numbers of instances depending on their availability. They are compensated 9 GBP/hour for their work. We paid 606.41 GBP in total.

They score each sample by following the rubric in \Cref{tab:rubric}. We engage evaluators by asking them to write at least 30 characters to describe their rationale for preference, which also helps evaluators take the rating more seriously and derive more rational and accurate ratings. We also provide an evaluation session with no more than evaluating 5 outlines considering the human attention span. If evaluators want to continue participating, they have to sign up for another evaluation session, which ensures they have a break for recharging themselves. Lastly, we utilize Prolific's offering that automatically rejects work that takes too long or too short, above or below two standard deviations of the average completion time.

\subsection{RAG Response Evaluation}
\label{sec:ragevaldetail}

\begin{table}[h]
\small
\renewcommand{\arraystretch}{1.2}
\begin{threeparttable}
\resizebox{\textwidth}{!}{%
\begin{tabular}{@{}rcc|rcc@{}}
\toprule
\multicolumn{3}{c|}{\textbf{Search Query Evaluation}} & \multicolumn{3}{c}{\textbf{Content Draft Evaluation}} \\ \midrule
& \multicolumn{2}{c|}{\textbf{DPO-\model}} & & \multicolumn{2}{c}{\textbf{DPO-\model}} \\
\textbf{Vanilla RAG} & \multicolumn{1}{c}{Unsatisfactory} & \multicolumn{1}{c|}{Satisfactory} & \textbf{SFT-\model} & \multicolumn{1}{c}{Unsatisfactory} & \multicolumn{1}{c}{Satisfactory} \\
Unsatisfactory       & 179                                & 606       & Unsatisfactory        & 130                                & 271      \\
Satisfactory         & 60                                 & 155                               & Satisfactory          & 207                                & 392                              \\ \bottomrule
\end{tabular}
}
\begin{tablenotes}
\small
\item Note: The sum of the counts in the contingency tables is 1000 (100 queries evaluated by 10 workers) for each evaluation, respectively. 
\end{tablenotes}
\end{threeparttable}
\caption{Contingency tables for human study in response evaluation}
\label{tab:contingency}
\end{table}

A total of 63 crowd workers participate in the evaluation (Gender: 33 men and 30 women; Age: Mean=28.08 yrs, SD=9.01 yrs, Min=19 yrs, Max=68 yrs; Ethnicity: White: 26, Black: 22, Mixed: 10, and Asian: 5; Country of residence: South Africa: 20 (15.87\%), Portugal: 10 (7.94\%), Mexico: 8 (6.35\%), Poland: 6 (4.76\%), Canada: 4 (3.17\%), and 11 other countries; Highest education level completed: A majority of the evaluators hold at least a Bachelor's degree (n=49, 77.78\%)). Individual crowd workers evaluate different numbers of instances depending on their availability. They are compensated 9 GBP/hour for their work. We pay 878.58 GBP in total.

We offer ten single sessions for evaluation (5 sessions for Vanilla RAG vs. DPO-\model and 5 sessions for SFT-\model vs. DPO-\model). Each session has ten evaluators. If wanted, evaluators can participate in more than one session; thirteen out of 63 evaluated multiple sessions. A session takes 20 to 40 minutes. In a session, evaluators are first provided with a short tutorial with evaluation guidelines and examples. Then, they evaluate a pair of responses that answer the same query for twenty queries. They are told that all of the queries are formatted as \texttt{[question]} ($q_{base}$) + \texttt{[instruction]} ($q_{cov}$) and asked to mark a response ``Satisfactory'' if the response satisfies any of the two evaluation items: (1) the response indeed answers the question \texttt{[question]} with evidence or partial evidence, which includes ``there is no evidence but here is useful information,'' and (2) the response follows the instruction, \texttt{[instruction]}, and mark it ``Unsatisfactory'' otherwise. If all of the two responses to the same query are rated ``Satisfactory,'' they were asked to choose which one was a better answer (155 cases in Search Query Evaluation and 392 cases in Response Outline Evaluation; refer to \Cref{tab:contingency}). We describe a better answer would have more items described above satisfied, or it would be better at following the instruction. 

\subsubsection{Document Retrieval}
\label{sec:retrieval}

For simulating RAG pipeline, we utilize DuckDuckgo\footnote{\url{https://serpapi.com/duckduckgo-search-api}} to search relevant documents. To balance the number of documents, top-10 documents are retrieved in Vanilla RAG, and top-2 documents are retrieved for each subquery (\ie, 2 * 5 = 10 documents including \cc query) in SFT-\model and DPO-\model. Additionally, we follow the \textit{associative selection} process, suggested in \citet{pmlr-v202-lee23n}, to extract relevant evidence paragraphs from retrieved documents. Specifically, we construct FLAN-T5-Large~\citep{JMLR:v25:23-0870} trained with Wikipedia-based datasets such as MS-MARCO~\citep{bajaj2018ms}, ELI5~\citep{fan-etal-2019-eli5}, ASQA~\citep{stelmakh-etal-2022-asqa}, and Qasper~\citep{dasigi-etal-2021-dataset}. The trained language model matches passages in each document with given subqueries and returns an answerability score deciding whether the paired subquery and passage are relevant. We select the top-1 passage for each document as evidence for generating RAG response.

\end{document}